%% file: main.tex
\documentclass[11pt]{article}
\usepackage{tikz}
\usepackage{pgf-pie}
\usepackage{xcolor}
\usepackage{acl}
\usepackage[utf8]{inputenc}
\usepackage{times}
\usepackage{latexsym}
\usepackage{multirow}
\usepackage{tikz}
\usepackage{booktabs}
\usepackage{pgfplots} \pgfplotsset{compat=1.17}
\usepackage{xcolor}  
\usepackage{pifont}
\usepackage{newunicodechar} 
\newunicodechar{↔}{\leftrightarrow}
\newcommand{\checkmark}{\textcolor{green}{\ding{51}}}
\usepackage[T1]{fontenc}
\usepackage[prependcaption,textsize=tiny]{todonotes}

\usepackage{microtype}

\title{Mitrasaṃgraha: A Comprehensive Classical Sanskrit Machine Translation Dataset}

\author{
  {\bf Sebastian Nehrdich}$^{1}$ \quad
  {\bf David Allport}$^{2}$ \quad
  {\bf Sven Sellmer}$^{3}$ \\
  {\bf Jivnesh Sandhan}$^{4}$ \quad
  {\bf Manoj Balaji Jagadeeshan}$^{5}$ \quad
  {\bf Pawan Goyal}$^{5}$ \quad
  {\bf Sujeet Kumar}$^{5,6}$ \quad
  {\bf Kurt Keutzer}$^{2}$
  \\
  \\
  $^1$Center for Integrated Japanese Studies, Tohoku University $^2$University of California, Berkeley \\
  $^3$Adam Mickiewicz University in Poznań $^4$Kyoto University \\
  $^5$Indian Institute of Technology, Kharagpur \\
  $^6$B. P. Mandal College of Engineering Madhepura \\
}

\begin{document}
\maketitle
\begin{abstract}
While machine translation is regarded as a "solved problem" for many high-resource languages, close analysis quickly reveals that this is not the case for content that shows challenges such as poetical language, philosophical concepts, multi-layered metaphorical expressions, and more. Sanskrit literature is a prime example of this, as it combines a large number of such challenges in addition to inherent linguistic features like sandhi, compounding, and heavy morphology, which further complicate NLP downstream tasks. It spans multiple millennia of text production time as well as a large breadth of different domains, ranging from ritual formulas via epic narratives, philosophical treatises, poetical verses up to scientific material. As of now, there is a strong lack of publicly available resources that cover these different domains and temporal layers of Sanskrit.\\
We therefore introduce Mitrasaṃgraha, a high-quality Sanskrit-to-English machine translation dataset consisting of 391,548 bitext pairs, more than four times larger than the largest previously available Sanskrit dataset Itihāsa~\cite{2021itihasa}. It covers a time period of more than three millennia and a broad range of historical Sanskrit domains. In contrast to web-crawled datasets, the temporal and domain annotation of this dataset enables fine-grained study of domain and time period effects on MT performance. We also release a validation set consisting of 5,587 and a test set consisting of 5,552 post-corrected bitext pairs. We conduct experiments benchmarking commercial and open models on this dataset and fine-tune NLLB and Gemma models on the dataset, showing significant improvements, while still recognizing significant challenges in the translation of complex compounds, philosophical concepts, and multi-layered metaphors. We also analyze how in-context learning on this dataset impacts the performance of commercial models.

\end{abstract}

\section{Introduction}
Sanskrit is one of humanity's longest-attested literary traditions, encompassing ritual hymns, epics, scientific treatises, philosophical commentaries, poetry and more, composed over a span of three millennia. Yet, despite its enduring impact on South Asian culture, Sanskrit remains a severely low-resource language for contemporary NLP. In machine translation (MT), existing parallel corpora are confined to narrow domains and periods, contain relatively few sentence pairs, and therefore fail to represent the topical breadth and temporal depth of Sanskrit literature. Consequently, Sanskrit-to-English MT, particularly for open, non-commercial systems, lags far behind the progress observed for high-resource modern languages, hindering both digital humanities research and public access to Sanskrit sources.\\
To address this gap we present Mitrasaṃgraha, a large-scale, carefully curated Sanskrit--English parallel corpus that spans from the early Vedic period until late-medieval scholastic works. Beyond assembling the data, we compare two sentence-alignment strategies tailored to noisy historical texts, create manually inspected development and test splits, validate automatic MT metrics against expert judgments, and benchmark both commercial LLMs and fine-tuned open-source models. We further provide a qualitative error analysis that highlights the specific linguistic phenomena such as compounding, free word order, meter, and polysemy, that make Sanskrit translation uniquely challenging.\\
This paper makes the following contributions:
\begin{itemize}
\item \textbf{Mitrasaṃgraha}, the largest public Sanskrit$\rightarrow$English MT corpus to date with 391,548 aligned sentence pairs plus 5,587-sentence development and 5,552-sentence test splits covering six literary domains and three millennia of Sanskrit.
\item A detailed \textbf{sentence-alignment study} on noisy historical texts; among four systems, we find \textsc{BertAlign} delivers the highest performance for Sanskrit--English.
\item Manually post-corrected \textbf{development/test sets} and \textbf{document-level metadata} (title, author, period, genre) released to support fine-grained evaluation and domain-transfer research.
\item An \textbf{empirical comparison of four automatic MT metrics}, where neural BLEURT and LLM-based GEMBA/GEMBA* show the strongest correlation with expert judgments, outperforming BLEU and chrF.
\item \textbf{Comprehensive MT baselines}: zero-shot commercial LLMs, fine-tuned open-source models, and retrieval-augmented generation setups, illustrating how the new dataset improves Sanskrit machine translation.
\end{itemize}
We make the dataset and fine-tuned model checkpoint available at \url{https://github.com/dharmamitra/mitrasamgraha-dataset}.\\
This paper is organized as follows. We first provide background on the Sanskrit language in Section~\ref{sec:sanskrit-lang} and review related work in Section~\ref{sec:related-work}. Section~\ref{sec:sentence-alignment} evaluates different sentence alignment approaches, while Section~\ref{sec:workflow} describes our data collection workflow. We present the Mitrasaṃgraha dataset in Section~\ref{sec:dataset}. We then evaluate machine translation metrics for Sanskrit in Section~\ref{sec:metrics}. We show benchmark experiments of different commercial and open-source models in Section~\ref{sec:benchmark-experiments}, followed by an analysis of translation challenges in Section~\ref{sec:analysis} and an ablation study on data cleaning and prompt augmentation in Section~\ref{sec:ablation-study}. Section~\ref{sec:conclusion} summarizes the paper. Section~\ref{sec:limitations} describes its limitations.

\subsection{The Sanskrit Language}\label{sec:sanskrit-lang}
Sanskrit is a classical language of the Indo-Aryan branch of the Indo-European languages. It was widely used as a lingua franca by religious, scientific and literary communities of ancient India. Since the second half of the first millennium BCE, Sanskrit was used for the production of Buddhist literature. Early Buddhist works in Sanskrit show a strong influence of Middle Indian dialects, which has decreased over time~\cite{bhsgr}. Sanskrit relies heavily on morphology to indicate grammatical relations and has a relatively free word order. It follows the nominative-accusative alignment, has a complex verbal system and a rich nominal declension. Nominal compounds are frequently found. Another special characteristic and challenge in the computational processing of Sanskrit is the phenomenon of sandhi, by which the contact phonemes of neighboring word tokens are changed and merged, and which creates unseparated strings spanning multiple tokens~\cite{hellwig2018}.

\section{Related Work}\label{sec:related-work}
For Classical Sanskrit, \citet{2021itihasa} introduced a dataset consisting of 93,000 pairs of Sanskrit verses with their English translations. This dataset consists of two central epic Sanskrit texts, the Mahābhārata and the Rāmāyaṇa, composed between the 4th century BCE and the 4th century CE. Both texts have been translated by the same person, M.N. Dutt in the late 19th century. 
Regarding contemporary Sanskrit, \citet{nllbteam2022language} provide 244k automatically mined sentence pairs. \citet{gala2023indictrans2} expand this dataset by adding 27.7k sentences from the Wiki and 5.4k sentences from the Daily domains (day-to-day spoken language). 
Another Sanskrit dataset was published in \citet{2024samayik}. This dataset consists of about 53k sentence pairs, which are written in contemporary prose and therefore do not cover the classical domain. When it comes to other language directions, \citet{nehrdich2022} provides a Sanskrit-Classical Tibetan sentence-aligned dataset that covers 317k sentence pairs.  
\input{input/dataset-comparision}

\section{Sentence Alignment For Sanskrit and English}\label{sec:sentence-alignment}
Aligning Sanskrit-English sentence pairs presents unique challenges compared to high-resource language pairs. Sanskrit's  sandhi (phonological changes at word boundaries), frequent compounding, and free word order weaken lexical alignment signals. Additionally, divergent punctuation conventions between Sanskrit editions and English translations create frequent one-to-many/many-to-one mappings, while English texts often contain extraneous material (footnotes, headers) with no Sanskrit counterpart.
We evaluated two state-of-the-art alignment algorithms on manually aligned gold-standard data with synthetic noise injection to mimic real-world conditions: \textsc{VecAlign} \citep{vecalign2019}, and \textsc{BertAlign} \citep{bertalign}. In line with findings on Sanskrit–Tibetan alignment by \citet{nehrdich2022}, we focus our benchmark on embedding-based aligners and omit purely length- and lexicon-based methods, which performed substantially worse in our setting. Following \citet{yasa}, we report sentence-level F-measure ($F_S$), which measures whether correct sentences are paired regardless of many-to-many mappings.
\textsc{BertAlign} demonstrated superior performance across all evaluated domains, achieving $F_S$ scores ranging from 70.15 (Tattvasaṃgraha) to 97.12 (Manusmṛti), significantly outperforming the other algorithms.  Based on these results, we adopted \textsc{BertAlign} for creating Mitrasaṃgraha. While certain domains still show room for improvement, the alignment quality is sufficient for downstream machine learning applications. Detailed alignment results and methodology are provided in Appendix~\ref{app:alignment}.

\section{Data Collection Workflow}\label{sec:workflow}
We follow a structured multi-step workflow for data collection and preprocessing/postprocessing. First, we collaborate with domain experts of Sanskrit literature to identify Sanskrit texts that have corresponding English translations. We then source these text pairs from publicly available resources.\\
For data extraction, we employ a systematic preprocessing pipeline. For texts available on public websites, we use automated scraping systems to extract high-quality content from HTML sources. For public domain PDFs, we apply an OCR pipeline using Google Cloud Vision, followed by manual cleanup to remove irrelevant sections such as introductions, indices, and notes. Once the text is preprocessed, we segment the Sanskrit source as well as the English translation into individual sentences using a rule-based approach with manual review, since depending on the input format, different rules need to be applied for successful sentence segmentation. These rules need to be selected and adjusted for each text individually. After the sentence segmentation, we utilize \texttt{BertAlign} to generate aligned bitext pairs. Once the automatic alignment is completed, we manually review the output and dismiss files with insufficient alignment quality. For those that we keep, we do a coarse manual inspection where we remove sections that are obviously misaligned.
\input{graphics/data-collection-workflow}

\section{Dataset Description}\label{sec:dataset}
We provide an overview of the domain composition of the dataset in Figure~\ref{fig:sanskrit_lit_distribution} and compare Mitrasaṃgraha with previously published datasets in Table~\ref{dataset-comparison}. A detailed per-document breakdown of the dataset is given in Appendix~\ref{app:dataset-train}. \\
To demonstrate the domain distribution, we assigned high-level categories to each text. These categories are intentionally broad; for instance, "treatises" encompasses Dharmaśāstra literature, medical texts, and commentaries on religious scriptures, among others. The temporal classification of Sanskrit literature presents significant challenges, as philological research typically establishes only approximate ranges, often spanning several centuries. Given these constraints, we divide our dataset into four broad periods: Vedic (before 500 BCE), Epic (500 BCE--200 CE), Classical (200--1000 CE), and Medieval (1000--1800 CE), while acknowledging the inherent imprecision of such categorization. \\
The Sanskrit sources in our dataset span nearly three millennia, from the second millennium BCE to the early eighteenth century CE. This breadth is complemented by the diversity of translators represented in the corpus, with at least forty identified contributors. Some works, such as the \textit{Mahāsubhāṣitasaṃgraha}, are anthologies comprising contributions from numerous additional translators. The combination of temporal range and domain variety means that our dataset captures the linguistic and stylistic evolution of Sanskrit literature across its major historical periods, from the earliest Vedic hymns to late medieval philosophical treatises.
\input{input/dataset-piechart}
\subsection{Compliance of Mitrasaṃgraha with ARR Guidelines for New Dataset Release}\label{sec:arr-compliance}
We release Mitrasaṃgraha in two versions to accommodate different licensing requirements: (1) a complete dataset under CC BY-NC-ND 4.0 containing all 391,548 sentence pairs, and (2) a research-friendly subset under CC BY-SA 4.0 containing 372,791 sentence pairs (excluding 18,757 pairs with BY-NC-ND restrictions). Detailed licensing information for each text is provided in Appendix~\ref{app:dataset-train}. 
All data collected or scraped from various sources (e.g., websites, PDF books) has been manually verified to be publicly accessible without copyright restrictions. The resources and pre-existing datasets incorporated into our collection are used in a manner consistent with their originally intended purposes (where specified) and are aligned with our declared intended use.
Our dataset does not raise any anonymity concerns, as it contains no personally identifiable information or offensive content.

\section{Machine Translation Metrics Evaluation for Sanskrit}\label{sec:metrics}
\label{sec:metrics}
 In order to assess the effectiveness of different automatic metrics for Sanskrit-to-English machine translation evaluation, we asked two expert annotators to independently assess machine-generated translations for 100 sentences. For each of the 100 sentences, we generated three translations with GPT-4o-mini, Claude 3.5 Haiku, and Claude 3.5 Sonnet. We then randomized the order of the translations, without disclosing the identity of the models to the annotators. We asked the annotators to determine for each sentence which of the three translations is the best, and which is the worst. All annotators either hold PhDs or are doctoral candidates specializing in Sanskrit literature.
 We conducted a one-hour annotation training session where the annotators together evaluated the machine translations of ten sentences, in order to ensure common evaluation standards. 
 We present the results for the inter-annotator agreement on determining the best and worst translations in Table~\ref{tab:inter-annotator-corr}. \\ 
 Evaluating Sanskrit-English machine translation presents unique challenges that impact the suitability of automatic metrics. Standard n-gram-based metrics such as BLEU and chrF are limited by specific features of Sanskrit. There is the phenomenon of sandhi, where phonemes at word boundaries change and merge, creating ambiguity in word segmentation. Second, Sanskrit has a relatively free word order compared to English, leading to variations in sentence structure that metrics struggle to capture. Additionally, Sanskrit literature often employs complex compounding, figurative language, and compressed verses, making direct surface-level comparison difficult. Finally, many core semantic concepts in philosophical or religious texts lack direct English equivalents, leading humans to adopt different valid approaches. All these factors lead to significant stylistic and structural variation even among high-quality human reference translations.\\  
To quantify the extent of this diversity in translation and its potential impact on evaluation, we conducted a focused analysis. We manually collected seven different high-quality human translations of 100 verses of the \textit{Bhagavad-gītā}. We then measured corpus-level BLEU, chrF, and reference-based BLEURT scores for all 42 possible directed pairwise comparisons among these seven human translations. The average pairwise BLEU score was 25.2, and the average pairwise chrF score was 44.0. This highlights the considerable surface-level divergence even among expert human renditions, showing that multiple perfectly fine English translations can substantially disagree when it comes to surface characteristics such as wording and structure.\\
 We investigate how different metrics correlate with expert human judgment in light of these challenges. 
 \input{input/inter-author-agreement}
In Figure~\ref{fig:avg-auto-vs-human} we show how the automatic metrics BLEU, chrF, BLEURT, and GEMBA perform against the human annotations. BLEU~\cite{bleu} and chrF~\cite{chrf} capture word n-gram precision/recall and character-level overlap and are calculated on document level. BLEURT~\cite{bleurt} uses a BERT model fine-tuned on human judgments of translation quality. We use the "BLEURT-20" checkpoint and average the score across all samples. GEMBA~\cite{gemba} uses LLMs to rate machine translation quality. We implement GEMBA using the Gemini 2.0 Flash API (April 2025). In our evaluation, GEMBA denotes the reference-based implementation, where we used the Sanskrit source sentence, the English reference translation, and the machine translation candidate for scoring. GEMBA* denotes the reference-free implementation which only uses the Sanskrit source sentence and the machine translation candidate for evaluation.\\ 
Our results show weak correlations for the string-based metrics BLEU and chrF, which is in line with previous research for high-resource languages~\cite{metrics-maze}, which established that these metrics are not suitable to compare machine translation quality across different model types. BLEURT, GEMBA*, and GEMBA all show significantly stronger performance. While the reference-based GEMBA slightly outperforms BLEURT and GEMBA*, it is noteworthy that the reference-free GEMBA* performs very competitively compared to the two reference-based neural models. This outcome is encouraging for the evaluation of domains where reference evaluation data is scarce. In summary, our results show that the neural model based score BLEURT and the LLM-based GEMBA*/GEMBA scores outperform string similarity-based metrics by a clear margin, making them the preferred choice for assessing Sanskrit-to-English machine translation quality across different model types.\\
 \input{input/metrics-agreement}

\section{Benchmark Experiments}\label{sec:benchmark-experiments}
In order to quantify the impact of Mitrasaṃgraha and to give an overview of the performance landscape of current MT systems for Sanskrit-to-English, we evaluate three families of translation systems.\\
Commercial, off-the-shelf LLM/MT APIs (zero-shot): Google Translate, GPT-4o-mini / GPT-4o (OpenAI), Claude 3.5 Haiku (1022) / Claude 3.5 Sonnet (0620) (Anthropic), and Gemini 1.5 Pro / Gemini 2.0 Flash (Google).\\
Open-source multilingual MT models, both in their original "zero-shot" form and after supervised fine-tuning on our training split: NLLB-200 600M \& 3.3B~\citep{nllbteam2022language}, and MADLAD-400-3B-MT~\citep{madlad400}. 
We also test the general-purpose instruction LLMs Llama-3-Instruct (8B)~\citep{llama3} and Gemma-2-Instruct (9B)~\citep{gemma2024} by prompting them as translators and by fine-tuning Gemma 2 9B on our corpus.\\
Retrieval-augmented generation (RAG): the top-performing commercial LLMs (Claude Sonnet, Gemini 1.5 Pro, Gemini 2.0 Flash) are paired with a sentence-level fine-tuned LaBSE retriever that feeds the most similar Sanskrit--English pairs from the training set into the prompt, following the recipe of~\citet{ram2023rag}.\\
All non-commercial models are decoded with greedy search. Performance is reported on the held-out test split using BLEU~\citep{papineni2002bleu}, chrF~\citep{popovic2015chrf}, and GEMBA~\citep{kocmi2023gemba}; the latter is the metric that showed the strongest correlation with expert judgments in Section~\ref{sec:metrics}. We show the results in Table~\ref{tab:translation-results}. The first group are the commercial baselines, the second group the commercial baselines retrieval-augmented with the Mitrasaṃgraha dataset, the third group the open model baselines, and the fourth group the open models after full-parameter fine-tuning on our dataset.\\
\input{input/results}

\section{Analysis}\label{sec:analysis}
We present the results of our comparative benchmark in Table~\ref{tab:translation-results}.\footnote{The results in this table were computed using an expanded training corpus that includes an additional augmentation set, increasing the size to a total of 516,000 sentence pairs for training. For licensing reasons, this augmentation material cannot be redistributed, and is therefore excluded from the publicly released version of the dataset. We report results on the expanded corpus to quantify the effect of larger-scale training, while expecting the qualitative trends and relative comparisons to remain stable under the redistributable release.} Regarding the results for the \textbf{commercial baselines}, it is evident that the machine translation provider Google Translate underperforms compared to all chat systems provided by OpenAI, Claude, and Google. The best performance is achieved by Claude 3.7 Sonnet in BLEURT and Gemini 2.0 Flash in GEMBA. All commercial models that have in-context learning abilities (which excludes Google Translate) benefit significantly from retrieval augmentation, with the effect most strongly pronounced for Claude 3.7 Sonnet.\\
When it comes to the open model baseline comparison, the decoder-only LLMs Llama-3.1-8B-Instruct and Gemma-2 9B significantly outperform the encoder-decoder models NLLB, MADLAD-400, and IndicTrans2. The encoder-decoder-only models without any fine-tuning struggle to achieve meaningful performance at all and frequently generate nonsensical output. As shown with NLLB, the higher parameter count is not leading to better baseline performance. The decoder-only LLMs are able to provide much better results, but they fall short compared to even the weakest commercial chat systems GPT-4o-mini and Claude 3.7 Haiku, while being somewhat comparable to Google Translate in chrF and BLEURT scores.\\
The open models all benefit significantly from full-parameter fine-tuning as shown in the last section of Table~\ref{tab:translation-results}. NLLB with 3.3B size outperforms NLLB 600M in all metrics. Gemma-2 9B fine-tuned significantly outperforms the encoder-decoder models here.\\


\section{Ablation Study}\label{sec:ablation-study}
To study the influence of noisy samples on model performance, we conduct an ablation study with different data cleaning methods. Given the alignment results shown in Table~\ref{aligner-eval}, we can expect noise to be present in our dataset. We therefore evaluate two approaches for removing noisy datapoints: (1) a deterministic rule-based method combining fixed criteria such as string length ratios with MT evaluation scores using multiple translation models as references; and (2) an LLM-based cleaning method utilizing a specialized prompt pattern.

For the deterministic cleaning approach, we begin with 391,548 datapoints and first filter out samples where the input-output string length ratio falls outside a predetermined range. We then train a Gemma-2-based machine translation model on the remaining dataset to serve as our initial reference. Each sample is evaluated against the MT output of this reference model using a combination of BLEU, chrF, and BERTScore metrics, identifying approximately 20\% of datapoints as potentially faulty.

For these flagged sentences, we generate additional reference translations using both Google Translate and Claude 3.5 Sonnet. We then compute evaluation scores (BLEU, chrF, and BERTScore) between the flagged datapoints and these MT/LLM-generated references. Samples with scores below a certain threshold for all three metrics across all three models (our fine-tuned Gemma-2, Google Translate, and Claude 3.5 Sonnet) are removed. This process yields a cleaned dataset of 368,415 datapoints. While using a model trained on the same data to detect faulty samples introduces potential bias, we mitigate this by incorporating external models as additional voters in the pruning process.\\
For the LLM-based approach, we process the complete set of 391,548 datapoints using Gemini 1.5 Pro, employing a specialized prompt pattern to clean and post-correct the alignments.\\
We present the results in Table \ref{ablation-cleaning}. The results show that our deterministic filtering leads to performance increases of +5.01 GEMBA and +0.52 chrF on the fine-tuned NLLB model, while BLEURT shows no improvement. For the RAG setup, the cleaned dataset does not perform significantly better on any metric. Adding grammatical analysis to the Gemini 1.5 Pro as a prompt augmentation technique yields slight performance increases. However, when combining grammatical analysis with the RAG setup, this significantly decreases performance compared to using the RAG setup alone. The dataset cleaned with Gemini 1.5 prompting visibly underperforms compared to both the uncleaned base dataset and the deterministically cleaned dataset across all metrics.
\input{input/results-ablation}

\section{Conclusion}\label{sec:conclusion}
In this paper, we introduced the creation and release of the novel, comprehensive Mitrasaṃgraha dataset for Sanskrit-to-English machine translation, consisting of 391,548 bitext pairs. This dataset is now the largest and most comprehensive publicly available Sanskrit-English parallel corpus, being more than four times larger than Itihāsa and covering multiple domains across Vedic, Epic, Classical, and Medieval time periods. Alongside this training corpus, we contribute manually verified validation and test sets, each comprising approximately 5,500 bitext pairs. Our work included a careful examination of automatic alignment methods, establishing BertAlign in combination with a fine-tuned LaBSE embedding model as the most reliable method for aligning Sanskrit-English sentences.\\
Our MT metrics evaluation compared different scores, confirming that neural BLEURT and the LLM-based GEMBA score correspond much more closely with human judgment compared to string-similarity-based metrics like BLEU and chrF. We conducted a comprehensive benchmark of various open and commercial models, showing that fine-tuning open models such as NLLB and Gemma~2 on Mitrasaṃgraha leads to consistent performance increases. Furthermore, we demonstrated that in-context learning, by retrieving relevant samples from this dataset, clearly enhances the performance of commercial models.\\
Despite these advances, we recognize that the complex features of Sanskrit such as sandhi, compounding, multi-layered metaphors, free word order, and more present a uniquely challenging case for machine translation. Current MT solutions still exhibit significant weaknesses in adequately capturing these features. The Mitrasaṃgraha dataset, therefore, offers a foundational resource for low-resource Sanskrit NLP, can aid digital humanities research on Sanskrit literature, and improve public access to Sanskrit cultural heritage.
\section{Limitations}\label{sec:limitations}
Several limitations should be acknowledged. First, the creation of Mitrasaṃgraha faced challenges due to frequent mismatches in sentence segmentation between Sanskrit sources and their English translations, leading to complex one-to-many or many-to-one alignments. While our chosen aligner, BertAlign, is designed for such scenarios, these alignments inherently carry a higher risk of error, potentially introducing noise. Additionally, as large portions of the dataset were sourced from OCR-processed texts, persistent OCR mistakes are likely present despite preprocessing.\\
Second, while Mitrasaṃgraha provides extensive parallel data, the dataset itself does not inherently solve deep semantic challenges in Sanskrit MT, such as the interpretation of multi-layered metaphors or complex philosophical concepts. Current models, even when trained on this dataset, might still struggle to adequately capture these nuanced linguistic phenomena.\\
Finally, although Mitrasaṃgraha significantly expands domain and temporal coverage, certain specialized domains within Sanskrit literature, such as technical treatises on astronomy or mathematics, remain entirely missing or are severely underrepresented. Addressing these gaps will require dedicated future data collection and curation efforts.
\bibliography{anthology,custom}
\appendix

\section{Detailed Sentence Alignment Evaluation}
\label{app:alignment}

\subsection{Evaluation Methodology}

Sentence alignment is considered largely solved for high-resource language pairs with matching punctuation and clean input, where even simple length-based algorithms perform reliably \citep{gale-church-1993-program, brown-etal-1991-aligning}. Sanskrit<>English, however, poses far greater challenges. Sanskrit editions and their English translations frequently diverge in sentence boundaries, omit or reorder passages, and the English side is often cluttered with footnotes, headers, and page numbers that survive OCR but have no Sanskrit counterpart. Divergent punctuation conventions produce many one-to-many or many-to-one mappings, and Sanskrit's pervasive sandhi and prolific compounding weaken the surface lexical cues that many alignment approaches implicitly benefit from.\\
Following \citet{yasa}, we report two complementary scores: the alignment-level F-measure ($F_A$), which counts a link as correct only when the entire bisegment exactly matches the gold reference, and the sentence-level F-measure ($F_S$), which checks whether the correct sentences are paired irrespective of how they are grouped into bisegments. Using both metrics captures performance under a strict criterion that penalizes many-to-many links ($F_A$) and a more permissive view that tolerates them ($F_S$). Because many-to-many alignments can harm $F_A$ while leaving $F_S$ relatively unaffected, both measures are necessary to characterize alignment quality.\\
We begin with manually aligned gold pairs and synthetically corrupt them to mimic realistic noise. Specifically, we (i) randomly split some source or target segments on punctuation, creating one-to-many / many-to-one cases, and (ii) inject random extraneous English sentences (footnotes, headers, etc.). This stresses the aligners' ability to match uneven sentence counts while discarding irrelevant material.

\input{input/alignment-benchmark}

We present the results of the alignment benchmark in Table \ref{aligner-eval}. Across datasets, the embedding-based aligners \textsc{VecAlign} and \textsc{BertAlign} perform strongly under the sentence-level metric $F_S$, indicating that they typically identify the correct sentence pairings even when boundaries are noisy. \textsc{BertAlign} generally outperforms \textsc{VecAlign} in both $F_A$ and $F_S$, suggesting it is the more reliable choice for Sanskrit--English alignment in this setting.

Even \textsc{BertAlign} attains relatively modest $F_A$ scores, which is expected given frequent many-to-many correspondences and boundary divergence between editions and translations. For downstream use, the $F_S$ scores are more indicative; they range from 70.15 to 97.12, suggesting accuracy sufficient for many machine learning applications. At the same time, the lower-performing domains (e.g., \textit{Tattvasaṃgraha} and \textit{Buddhacarita}) highlight that alignment quality remains uneven and warrants further attention in future work.

\section{Dataset Catalog}
\label{app:dataset-train}
\begin{table*}[t]
\centering
\small

\label{tab:corpus_catalog}
\begin{tabular}{@{}p{0.22\textwidth}p{0.10\textwidth}p{0.15\textwidth}p{0.08\textwidth}p{0.10\textwidth}p{0.10\textwidth}r@{}}
\toprule
\textbf{Title} & \textbf{Date} & \textbf{Translator} & \textbf{Period} & \textbf{Category} & \textbf{License} & \textbf{Pairs} \\
\midrule
Abhidharmakośabhāṣya & 350-450 & Pruden & Classical & treatises & cc0 & 10,827 \\
Abhidharmakośakārikā & 350-450 & Pruden & Classical & treatises & cc0 & 597 \\
Agni-Purāṇa & 700-1000 & Ganghadharan & Classical & purana & BY-NC-ND 3.0 & 14,207 \\
Āpastamba-Dharmasūtra & 600BCE-300BCE & Bühler & Vedic & vedic & cc0 & 307 \\
Arthaśāstra & 300BCE-100BCE & Shama Sastry & Epic & treatises & cc0 & 5,346 \\
Arthaviniścayasūtra & 100-800 & Ānandajoti & Epic & rel\_script & cc0 & 5,052 \\
Atharvaveda-Saṃhitā & 1200BCE-1000BCE & Griffiths & Vedic & vedic & cc0 & 2,822 \\
Bhāmatī & 850-900 & Sastri Raja & Classical & treatises & cc0 & 1,350 \\
Bhāvanākrama & 770-780 & Sharma & Classical & treatises & cc0 & 873 \\
Chāndogya-Upaniṣad (Śaṅkarabhāṣya) & 700-900 & Jha & Classical & treatises & cc0 & 5,109 \\
Dhammapada & 100-300 & Ānandajoti & Epic & rel\_script & cc0 & 842 \\
Garga Saṃhitā & 1000-1300 & Goswami & Medieval & rel\_script & cc0 & 4,436 \\
Gautama-Dharmasūtra & 600BCE-400BCE & Bühler & Vedic & vedic & cc0 & 518 \\
Harivaṃśa & 200-500 & Dutt & Epic & epic & cc0 & 7,468 \\
Jātakamālā & 400-500 & Speyer & Classical & treatises & cc0 & 2,683 \\
Jīvanmuktiviveka & 1350-1400 & Swamy & Medieval & rel\_script & cc0 & 1,756 \\
Kathāsaritsāgara & 1000-1100 & Tawney & Medieval & poetry & cc0 & 12,515 \\
Kāvyādarśa & 500-700 & S K Belvalkar & Classical & poetry & cc0 & 417 \\
Kṛṣṇa-Yajurveda & 1200BCE-1000BCE & Keith & Vedic & vedic & cc0 & 7,501 \\
Kumārasambhava & 350-450 & Kale & Classical & poetry & cc0 & 424 \\
Madhyāntavibhāgabhāṣya & 350-450 & Anacker & Classical & treatises & cc0 & 630 \\
Madhyāntavibhāgaṭīkā & 550-600 & Stanley & Classical & treatises & cc0 & 2,601 \\
Madhyāntavibhāgaṭīkā 1 & 550-600 & Shcherbatskoy & Classical & treatises & cc0 & 888 \\
Mahābhārata & 400BCE-400CE & Ganguly & Epic & epic & cc0 & 90,789 \\
Manubhāṣya & 800-900 & Jha & Classical & treatises & cc0 & 19,919 \\
Manusmṛti & 200BCE-200CE & Bühler & Epic & rel\_script & cc0 & 1,384 \\
Meghadūta & 350-450 & Kale & Classical & poetry & cc0 & 103 \\
Mīmāṃsāsūtra Tantravarttika & 650-750 & Jha & Classical & treatises & cc0 & 6,171 \\
Nyāyamañjari & 900-1000 & Bhattacharyya & Classical & treatises & cc0 & 7,763 \\
Pañcaskandhaprakaraṇa & 350-450 & Anacker & Classical & treatises & cc0 & 151 \\
Raghuvaṃśa & 350-450 & Kale & Classical & poetry & cc0 & 488 \\
Rigveda & 1800BCE-1200BCE & Oldenberg & Vedic & vedic & cc0 & 11,645 \\
Sāmaveda-Saṃhitā & 1200-1000BCE & Griffths & Vedic & vedic & cc0 & 1,343 \\
Sāṃkhyakārikā & 300-400 & Horace Hayman Wilson & Classical & treatises & cc0 & 606 \\
Sarva-darśana-saṃgraha & 1370-1380 & Cowell & Medieval & treatises & cc0 & 1,755 \\
Śatapatha-Brāhmaṇa & 800-699BCE & Eggeling & Vedic & vedic & cc0 & 6,193 \\
Śiva-Purāṇa & 750-1200 & Shastri & Classical & purana & cc0 & 19,136 \\
Skanda-Purāṇa & 700-1200 & Tagare & Classical & purana & cc0 & 67,176 \\
Śukla-Yajurveda & 1100BCE-900BCE & Griffiths & Vedic & vedic & cc0 & 1,776 \\
Suśrutasaṃhitā & 200-500 & Bhishagratna & Classical & treatises & cc0 & 10,233 \\
Tantrāloka & 800-1100 & Singh & Classical & rel\_script & BY-NC-ND & 4,550 \\
Tattvasaṃgraha & 760-780 & Jha & Classical & treatises & cc0 & 981 \\
Tattvasaṃgraha Pañjikā & 780-790 & Jha & Classical & treatises & cc0 & 17,186 \\
Triṃśikāvijñaptibhāṣya & 550-600 & Jacobi & Classical & treatises & cc0 & 635 \\
Triṃśikāvijñaptikārikā & 350-400 & Anacker & Classical & treatises & cc0 & 20 \\
Viṃśatikā & 350-450 & Anacker & Classical & treatises & cc0 & 162 \\
Viṣṇu-Purāṇa & 400-700 & Horace Hayman Wilson & Classical & purana & cc0 & 3,574 \\
Viṣṇusmṛti & 300-600 & Jolly & Classical & rel\_script & cc0 & 2,831 \\
Yogavasiṣṭha & 900-1300 & Mitra & Classical & rel\_script & cc0 & 25,801 \\
\midrule
\multicolumn{6}{r}{\textbf{Total sentence pairs:}} & \textbf{391,548} \\
\bottomrule
\end{tabular}
\caption{Complete catalog of the documents contained in Mitrasaṃgraha with metadata and sentence pair counts.}
\end{table*}
\end{document}

%% file: input/dataset-comparision.tex
\begin{table*}[h]
\centering
\begin{tabular}{l|c|c|c|c|c}
\textbf{Metric} & \textbf{Itihasa} & \textbf{NLLB} & \textbf{BPCC} & \textbf{Sāmayik} & \textbf{Mitrasaṃgraha} \\
\hline
Released & 2021 & 2022 & 2022 & 2023 & 2026 \\
\hline
Domains & 1 & 1 & 1 & 5 & 6 \\
\hline
Sanskrit composition time & Epic & Modern & Modern & Modern & \begin{tabular}[c]{@{}c@{}}Vedic, Epic,\\Classical, Medieval\end{tabular} \\

\hline
Number of Translators & 1 & unknown & unknown & unknown & 40+ \\
\hline
Number of Datapoints & 93,000 & 244,367 & 277,467 & 53,000 & 391,548 \\
\hline
Document-level & \textcolor{red}{\ding{55}} & \textcolor{red}{\ding{55}} & \textcolor{red}{\ding{55}} & \textcolor{red}{\ding{55}} & \textcolor{green}{\checkmark} \\
\hline
Contains Prose & \textcolor{red}{\ding{55}} & \textcolor{red}{\ding{55}} & \textcolor{red}{\ding{55}} & \textcolor{red}{\ding{55}} & \textcolor{green}{\checkmark} \\
\hline
Contains Metrical & \textcolor{green}{\checkmark} & \textcolor{red}{\ding{55}} & \textcolor{red}{\ding{55}} & \textcolor{red}{\ding{55}} & \textcolor{green}{\checkmark} \\
\end{tabular}
\caption{Comparison of Mitrasamgraha with existing published Sanskrit-English datasets. }
\label{dataset-comparison}
\end{table*}

%% file: graphics/data-collection-workflow.tex
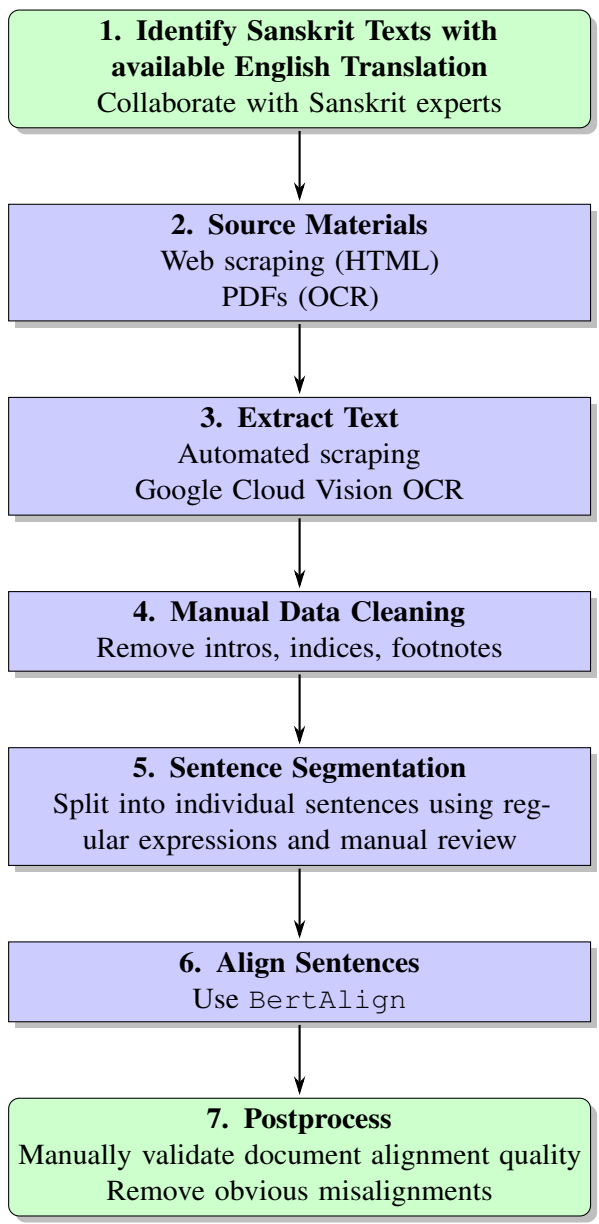
\begin{figure}[h]
  \centering
  \usetikzlibrary{shadows,arrows.meta,positioning}
  \begin{tikzpicture}[
    >=Stealth,                      
    startstop/.style={
      rectangle, rounded corners, draw=black, fill=green!20,
      text width=7.4cm, text centered, minimum height=1cm, drop shadow},
    process/.style={
      rectangle, draw=black, fill=blue!20,
      text width=7.4cm, text centered, minimum height=1cm, drop shadow},
    arrow/.style={
      draw=black, -{Stealth[length=2.5mm,width=1.5mm]},
      line width=0.8pt, shorten >=1pt, shorten <=1pt},
    node distance=1cm and 2cm
  ]
    \node[startstop] (identify) {\textbf{1. Identify Sanskrit Texts with available English Translation}\\Collaborate with Sanskrit experts};
    \node[process, below=of identify] (source) {\textbf{2. Source Materials}\\Web scraping (HTML) \\PDFs (OCR)};
    \node[process, below=of source] (extract) {\textbf{3. Extract Text}\\Automated scraping \\Google Cloud Vision OCR};
    \node[process, below=of extract] (cleanup) {\textbf{4. Manual Data Cleaning}\\Remove intros, indices, footnotes };
    \node[process, below=of cleanup] (segment) {\textbf{5. Sentence Segmentation}\\Split into individual sentences using regular expressions and manual review};
    \node[process, below=of segment] (align) {\textbf{6. Align Sentences}\\Use \texttt{BertAlign}};
    \node[startstop, below=of align] (post) {\textbf{7. Postprocess}\\Manually validate document alignment quality\\Remove obvious misalignments};

    \draw[arrow] (identify) -- (source);
    \draw[arrow] (source) -- (extract);
    \draw[arrow] (extract) -- (cleanup);
    \draw[arrow] (cleanup) -- (segment);
    \draw[arrow] (segment) -- (align);
    \draw[arrow] (align) -- (post);
  \end{tikzpicture}
  \caption{Flowchart of our Sanskrit-English bitext collection and alignment procedure.}
  \label{fig:data-workflow}
\end{figure}

%% file: input/dataset-piechart.tex
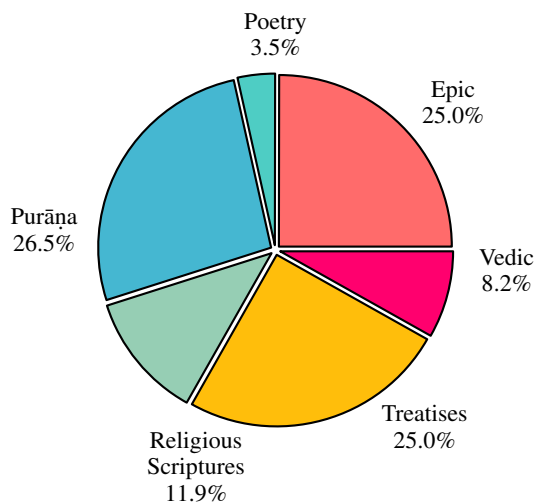
\begin{figure}[t]
\centering
\begin{tikzpicture}[scale=0.65]
\definecolor{epicred}{RGB}{255,107,107}    
\definecolor{poetryturquoise}{RGB}{78,205,196}    
\definecolor{puranaBlue}{RGB}{69,183,209}    
\definecolor{religiousGreen}{RGB}{150,206,180}    
\definecolor{treatisesYellow}{RGB}{255,190,11}    
\definecolor{vedicPink}{RGB}{255,0,110}    

\pie[hide number,
    radius=3.5,
    color={
        epicred,
        poetryturquoise,
        puranaBlue,
        religiousGreen,
        treatisesYellow,
        vedicPink            
    },
    text=white,
    sum=auto,
    after number=\%,
    rotate=0,
    explode=0.1,
    pos={0,0}
]{
    25.0,
    3.5,
    26.5,
    11.9,
    25.0,
    8.2
}

\node[font=\footnotesize, align=center] at (40:4.7) {Epic\\25.0\%};
\node[font=\footnotesize, align=center] at (90:4.4) {Poetry\\3.5\%};
\node[font=\footnotesize, align=center] at (-185:4.7) {Purāṇa\\26.5\%};
\node[font=\footnotesize, align=center] at (-110:4.7) {Religious\\Scriptures\\11.9\%};
\node[font=\footnotesize, align=center] at (-50:4.7) {Treatises\\25.0\%};
\node[font=\footnotesize, align=center] at (-5:4.7) {Vedic\\8.2\%};

\end{tikzpicture}
\caption{Distribution of Sanskrit Literature Categories}
\label{fig:sanskrit_lit_distribution}
\end{figure}

%% file: input/inter-author-agreement.tex
\begin{table}[t]
\centering
\begin{tabular}{lccc}
\toprule
\textbf{Measure} & \textbf{Best} & \textbf{Worst} & \textbf{Avg} \\
\midrule
Pearson’s \(\rho\)  & 0.549 & 0.554 & 0.552 \\
Spearman’s \(\rho\) & 0.544 & 0.551 & 0.548 \\
\bottomrule
\end{tabular}
\caption{Inter‑annotator correlation between Annotator A and Annotator B on “best” vs. “worst” translation rankings, with the average of both.}
\label{tab:inter-annotator-corr}
\end{table}

%% file: input/metrics-agreement.tex
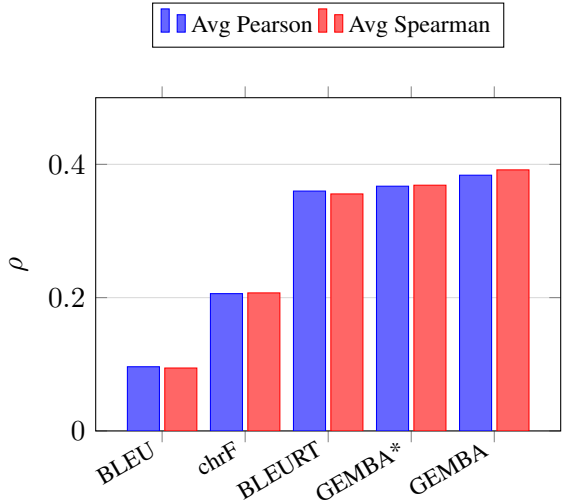
\begin{figure}[t]
  \centering
  \begin{tikzpicture}
    \begin{axis}[
      ybar,
      bar width=12pt,
      width=\linewidth,
      height=6cm,
      enlarge x limits=0.2,
      ylabel={\(\rho\)},
      ymin=0, ymax=0.5,
      ymajorgrids,
      grid style={gray!30},
      symbolic x coords={BLEU,chrF,BLEURT,GEMBA*,GEMBA},
      xtick=data,
      xticklabel style={rotate=30,anchor=east,font=\small},
      legend style={at={(0.5,1.15)},anchor=south,legend columns=2,font=\small}
    ]

    \addplot+[fill=blue!60] coordinates {
      (BLEU,0.0963) (chrF,0.2060) (BLEURT,0.3599)
      (GEMBA*,0.3672) (GEMBA,0.3837)
    };

    \addplot+[fill=red!60] coordinates {
      (BLEU,0.0944) (chrF,0.2071) (BLEURT,0.3556)
      (GEMBA*,0.3687) (GEMBA,0.3917)
    };

    \legend{Avg Pearson, Avg Spearman}
    \end{axis}
  \end{tikzpicture}
  \caption{Average Pearson (blue) and Spearman (red) correlations of automatic metrics vs. human judgments. GEMBA* is the reference-free implementation of GEMBA \cite{gemba}, while GEMBA is reference-based.}
  \label{fig:avg-auto-vs-human}
\end{figure}

%% file: input/results.tex
\begin{table*}[tbh]
\centering
\begin{tabular}{@{}l|lccc@{}}
\toprule
 & \textbf{Model} & \textbf{chrF} & \textbf{BLEURT} & \textbf{GEMBA} \\
\midrule
\multirow{7}{*}{\begin{tabular}[c]{@{}l@{}}Commercial\\Baselines\end{tabular}} 
 & Google Translate & 28.28 & 47.42 & 56.09 \\
 & GPT4o-mini & 29.44 & 50.16 & 66.39 \\ 
 & GPT4o & 32.63 & 53.53 & 82.37 \\
 & Claude 3.5 Haiku & 31.34 & 52.45 & 79.10 \\
 & Claude 3.7 Sonnet & 33.42 & \textbf{55.28} & 86.24 \\
 & Gemini 1.5 Pro & 31.62 &  51.20 & 79.61 \\
 & Gmini 2.0 Flash & \textbf{34.24} & 53.87 & \textbf{87.04} \\  
\midrule
\multirow{3}{*}{\begin{tabular}[c]{@{}l@{}}Commercial\\+ RAG\end{tabular}}
 & Claude 3.7 Sonnet & \textbf{38.83} & \textbf{57.54} & \textbf{89.32} \\ 
 & Gemini 1.5 Pro & 37.07 & 55.46 & 84.70 \\
 & Gemini 2.0 Flash & 38.80 & 56.69 & 88.61 \\ 
\midrule 

\multirow{5}{*}{\begin{tabular}[c]{@{}l@{}}Open Model\\Baselines\end{tabular}}
 & NLLB 600M  & 7.65 & 33.84 & 8.08 \\ 
 & NLLB 3.3B  & 1.43 & 10.07 & 0.82 \\
 & MADlad400-3b-mt  & 12.89 & 33.04 & 5.19 \\
 & IndicTrans2 1B & 16.76 & 39.35 & 12.23 \\ 
 & Llama-3.1-8B-Instruct  & 24.35 & 43.53 & 33.70 \\ 
 & Gemma2 9b it  & \textbf{27.58} & 47.97 & 50.64 \\ 
 & Gemma3 12b it & 25.78 & 47.92 & 55.68 \\
 & Gemma3 27b it & 27.5 & \textbf{49.13} & \textbf{64.55} \\ 
\midrule
\multirow{4}{*}{\begin{tabular}[c]{@{}l@{}}Open Model\\Finetuned\end{tabular}}
 & NLLB 600M & 35.59 & 52.88 & 71.07 \\
 & NLLB 3.3B & 36.06 & 53.89 & 75.12 \\
 & Gemma2 9b & \textbf{38.36} & \textbf{55.10} & \textbf{83.34} \\
\bottomrule
\end{tabular}
\caption{Translation performance across different models. Values shown are chrF, BLEURT and GEMBA scores on the test set. For each section, best scores are in \textbf{bold}.}
\label{tab:translation-results}
\end{table*}

%% file: input/results-ablation.tex
\begin{table}[tbh]
\footnotesize
    \centering
    \setlength{\tabcolsep}{4pt}
    \begin{tabular}{l|cccc}
    \toprule  
    Model  &  chrF & BLEURT & GEMBA  \\ 
    \hline
NLLB 600M ft (unclean) & 35.59 & \textbf{52.88} & 71.07 \\ 
NLLB 600M ft (clean) & \textbf{36.11} & 52.85 & \textbf{76.08} \\
\hline
G1.5 Pro Base & 31.62 & 51.20 & 79.61 \\ 
G1.5-Pro + Grammar & 32.42 & 51.52 & 81.07 \\
G1.5-Pro + RAG & \textbf{37.07} & 55.46 & \textbf{84.70} \\
G1.5-Pro + RAG + Gram & 34.57 & 54.01 & 84.66 \\
G1.5-Pro + RAG (clean) & 36.94 & \textbf{55.52} & 84.49 \\
G1.5-Pro + RAG (G-clean) & 35.81 & 54.83 & 84.13 \\ 
         \bottomrule
    \end{tabular}
    \caption{\label{ablation-cleaning}{Results of the ablation study on different data cleaning strategies and augmentation via grammatical annotation.}}
\end{table}

%% file: input/alignment-benchmark.tex
\begin{table}[tbh]
\centering
\resizebox{\columnwidth}{!}{%
\begin{tabular}{l|c|c|c|c}
\toprule
\multicolumn{1}{c}{} & \multicolumn{2}{c|}{VecAlign} & \multicolumn{2}{c}{BertAlign} \\
\hline
Dataset & $F_A$ & $F_S$ & $F_A$ & $F_S$ \\
\hline
Manusmṛti  & 60.64 & 82.67 & \textbf{82.88} & \textbf{97.12} \\
Tattvasaṃgraha & 15.11 & 63.55 & \textbf{30.00} & \textbf{70.15} \\
Buddhacarita & 63.07 & 79.55 & \textbf{85.50} & 74.62 \\
Itihasa Test  & 33.54 & 78.66 & \textbf{40.44} & \textbf{83.39} \\
Bhāgavata-Purāṇa & 64.41 & 74.79 & \textbf{72.63} & \textbf{85.05} \\
\bottomrule
\end{tabular}}
\caption{Comparison of VecAlign and BertAlign scores across datasets.}
\label{aligner-eval}
\end{table}